\documentclass[sigconf]{acmart}

\AtBeginDocument{%
  \providecommand\BibTeX{{%
    \normalfont B\kern-0.5em{\scshape i\kern-0.25em b}\kern-0.8em\TeX}}}

\setcopyright{acmcopyright}
\copyrightyear{2019}
\acmYear{2019}
\acmDOI{10.1145/1122445.1122456}

\usepackage{multirow}
\usepackage{hyperref}

\begin{document}

\title{Towards an Efficient ML System: Unveiling a Trade-off between Task Accuracy and Engineering Efficiency \\ in a Large-scale Car Sharing Platform}




\author{Kyung Ho Park*}
\email{kp@socar.kr}
\affiliation{%
  \institution{SOCAR AI Research}
  \city{Seoul}
  \country{Republic of Korea}
}

\author{Hyunhee Chung*}
\email{esther@socar.kr}
\affiliation{%
  \institution{SOCAR AI Research}
  \city{Seoul}
  \country{Republic of Korea}
}

\author{Soonwoo Kwon}
\email{tigger@socar.kr}
\affiliation{%
  \institution{SOCAR AI Research}
  \city{Seoul}
  \country{Republic of Korea}
}

\renewcommand{\shortauthors}{K.H. Park, H. Chung, and S. Kwon}


\begin{abstract}

\textcolor{black}{Upon the significant performance of the supervised deep neural networks, conventional procedures of developing ML system are \textit{task-centric}, which aims to maximize the task accuracy. However, we scrutinized this \textit{task-centric} ML system lacks in engineering efficiency when the ML practitioners solve multiple tasks in their domain. To resolve this problem, we propose an \textit{efficiency-centric} ML system that concatenates numerous datasets, classifiers, out-of-distribution detectors, and prediction tables existing in the practitioners' domain into a single ML pipeline. Under various image recognition tasks in the real world car-sharing platform, our study illustrates how we established the proposed system and lessons learned from this journey as follows. First, the proposed ML system accomplishes supreme engineering efficiency while achieving a competitive task accuracy. Moreover, compared to the \textit{task-centric} paradigm, we discovered that the \textit{efficiency-centric} ML system yields satisfactory prediction results on multi-labelable samples, which frequently exist in the real world. We analyze these benefits derived from the representation power, which learned broader label spaces from the concatenated dataset. Last but not least, our study elaborated how we deployed this \textit{efficiency-centric} ML system is deployed in the real world live cloud environment. Based on the proposed analogies, we highly expect that ML practitioners can utilize our study to elevate engineering efficiency in their domain.}

\end{abstract}






\maketitle

\section{Introduction} \label{sec:intro}
\begin{figure*}[htb!]
\centering
\begin{tabular} {c c} \\
\includegraphics[width=0.42\textwidth, trim={0cm 0cm 0cm 0cm}, clip]{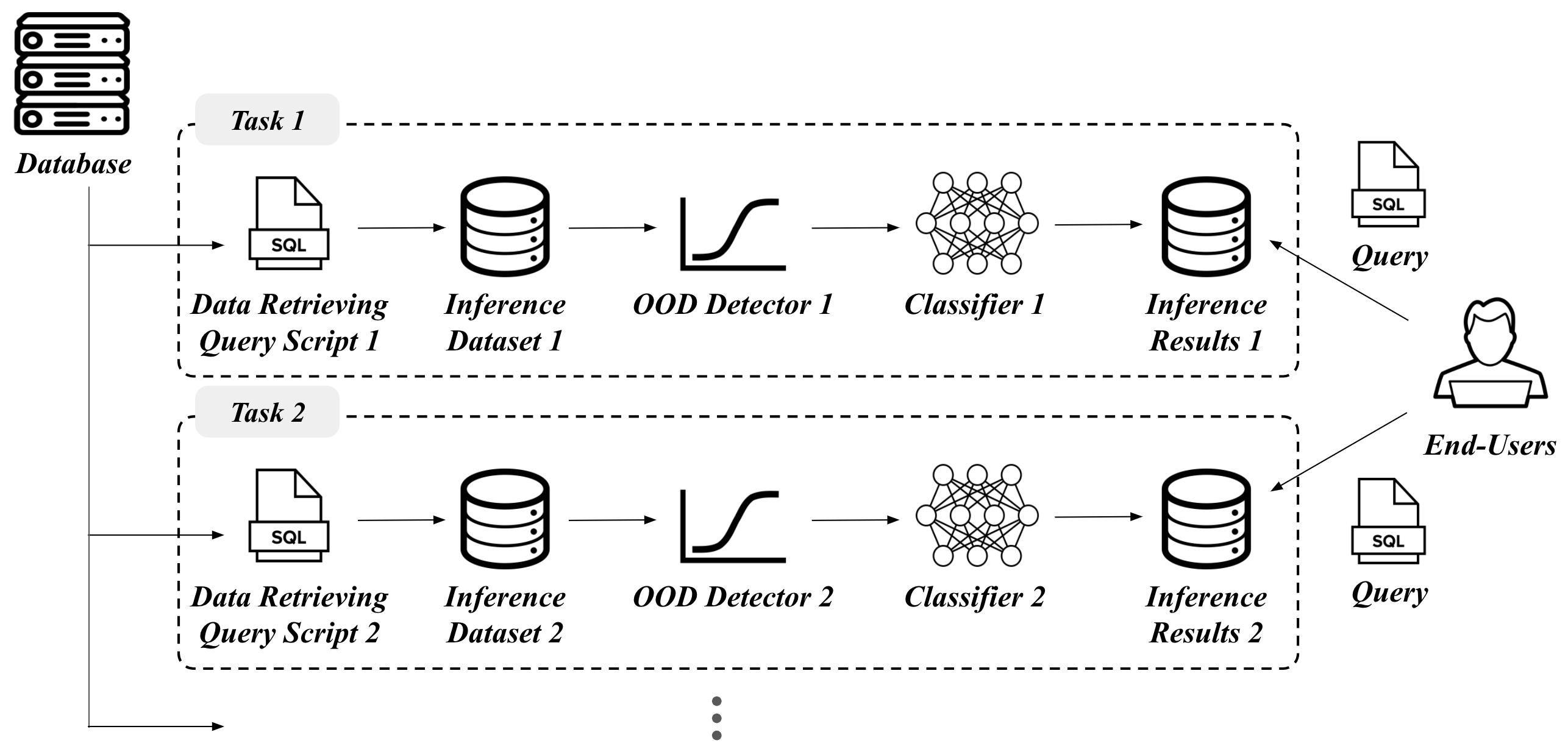} &
\includegraphics[width=0.42\textwidth, trim={0cm 0cm 0cm 0cm}, clip]{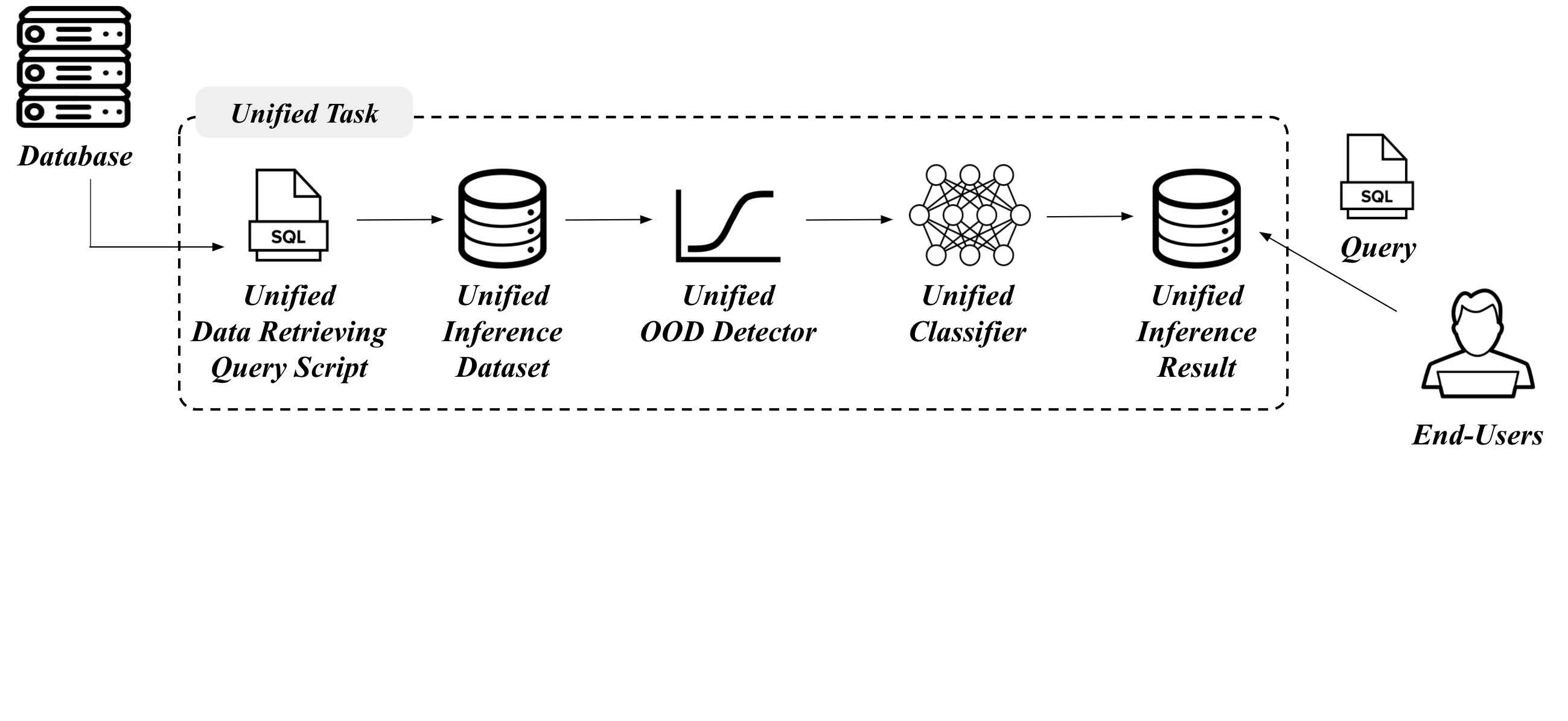} \\
\vspace{-0.1cm}
{\small (a) Task-Centric Approach } & {\small (b) Efficiency-centric Approach } \\
\end{tabular}
\caption{A comparison between the prior \textit{Task-centric} ML system and the proposed \textit{Efficiency-centric} ML system. The \textit{task-centric} ML system requires the practitioners to establish data-retrieving query scripts, OOD detector, classifier, and prediction result tables on each task. The end-user should refer to several independent tables to scrutinize a multi-labelable sample. On the other hand, our \textit{efficiency-centric} ML system resolves the aforementioned limits by concatenating the aforementioned components in the ML pipeline. The end-users under the \textit{efficiency-centric} paradigm can easily inspect multi-labelable samples in a single prediction table.}
\label{fig:approaches}
\end{figure*}

Recent progress of machine learning (ML) and convolutional neural networks (CNN) has empowered a great success in various real-world image recognition tasks \cite{nath2014survey}. When machine learning practitioners desire to solve a classification task in the real world, as shown in Figure. \ref{fig:approaches}. (a), conventional procedures of establishing an ML system are as follows. First, the practitioners build a labeling scheme and acquire a large amount of finely-labeled dataset. Second, they search for a state-of-the-art model fit to their task, implement the model, and train the classifier with the acquired dataset. Third, the practitioners build an out-of-distribution (OOD) detector if the model's performance is sufficient to be deployed. As the samples irrelevant to the task are often provided to the deployed classifier, the practitioners utilize the OOD detector to filter them out. Fourth, they deploy the trained classifier and OOD detector at the server. Lastly, they build an ML pipeline that sequentially performs the following steps. A system retrieves task-related data from the database with a query script, morph them as an inference dataset (which are candidate samples to the inference), filter out irrelevant samples with the OOD detector, provide the filtered inference data to the trained classifier, and accumulate the inference results to a prediction result table. When the pipeline is established, end-users (i.e., business operators, data Analysts) use the prediction result tables. When the ML practitioners encounter another task, they repeat the aforementioned procedures to solve a new task. In our study, we denote these procedures as a \textit{\textbf{task-centric approach}} as it focuses on solving an individual task with precise task performance.

However, the \textit{task-centric} approach bears several inefficiencies (denoted as \textit{engineering inefficiency} in our study) when the ML practitioners encounter multiple tasks. In real-world circumstances, ML practitioners, especially in the car sharing platform, usually encounter a particular class that is included in multiple tasks. Suppose the ML practitioners solve two classification tasks of \textit{label 0} v. \textit{label 1} and \textit{label 0} v. \textit{label 2}. Following the aforementioned \textit{task-centric} approach on two tasks, we analyzed there exists four primary inefficient points. First, in the perspective of model training, they should train the model twice for each task while the two tasks simultaneously utilize the \textit{label 0} class. At this point, we postulate a research question: "What if we concatenate two datasets (\textit{label 0} v. \textit{label 1} and \textit{label 0} v. \textit{label 2}) as a single one (\textit{label 0} v. \textit{label 1} v. \textit{label 2})? If a model trained with the concatenated dataset does not experience harsh performance degradation, can't we substitute two binary classifiers with a single 3-class classifier?" Second, there also exists room for improvement at the OOD detector. Establishing an effective OOD detector in the real world is challenging as the OOD detector cannot learn the irrelevant samples a priori \cite{hendrycks2018deep,hendrycks2016baseline}. Upon the difficulty of OOD detection, we posit another question: "If the concatenated dataset exists, would the OOD detector based on the concatenated dataset reject irrelevant samples better?" As the concatenated dataset let the model to learn wider knowledge compared to the \textit{task-centric} approach, we presume the OOD detector based on the concatenated dataset can improve the OOD detection performance. Third, in the perspective of deployment, \textit{task-centric} approach wastes a computing resource in a particular manner. Referring to two classifiers of \textit{label 0} v. \textit{label 1} and \textit{label 0} v. \textit{label 2}, the \textit{label 0} samples are included in a inference batch twice at each task, although the two trained classifiers would yield the same prediction of \textit{label 0}. Lastly, there also exists inconvenience for end-users when a sample is multi-labelable. Suppose a sample has attributes of both \textit{label 1} and \textit{label 2}. The \textit{label 0 v. label 1} classifier would yield a prediction of \textit{label 1}, and the \textit{label 0 v. label 2} classifier would yield a prediction of \textit{label 2}. As these results are accumulated in a separated prediction result tables, the end-users have to merge these two tables when they use these results in business operations. At this point, we also postulate the last question: "What if a single classifier trained on the concatenated dataset yields predictions of \textit{label 1} and \textit{label 2} simultaneously? If so, can't we just accumulate top-2 predictions at a single prediction result table and let the end-users use it?"

To resolve the proposed inefficiencies of the \textit{task-centric} approach, MLOps \cite{alla2021mlops,tamburri2020sustainable} paradigm utilizes various engineering tools to ease the task management with dataset versioning, model versioning, and fast model deployment. In other viewpoint, our study proposes a novel ML system paradigm denoted as \textit{\textbf{efficiency-centric}} approach, which resolves the inefficiencies of \textit{task-centric} approach by concatenating every task-centric dataset, model, OOD detector, and prediction result table into a single one. Based on two real-world image recognition tasks at SOCAR, the largest car-sharing platform in the Republic of Korea (it does operations similar to ZipCar in the United States), we introduced a series of analyses illustrating how we substituted a prior task-centric ML system into the \textit{efficiency-centric} ML system. To examine the effectivness of efficiency-centric paradigm, we postulate three research questions: 1) Does the concatenated classifier solve each task better?, 2) Does the concatenated OOD detector rejects irrelevant samples better? 3) Does the concatenated classifier calibrate well enough to provide useful prediction results on a multi-labelable sample? Throughout the study, we provide answers to these research questions and describe how we deployed the concatenated ML pipeline in the real world. We expect the proposed lessons learned will be a solid benchmark for candidate ML practitioners who desire to reduce inefficiencies at \textit{task-centric} paradigm. We acknowledge that this work is an applicative study in a large-scale car-sharing platform domain, and it aims to provide practical guidelines to machine learning practitioners who face similar challenges with us.

Throughout the study, key contributions are as follows:

\begin{itemize}
    
    \item We discovered the concatenated classifier accomplished a promising image recognition performance compared to the task-centric classifiers. However, as the concatenated classifier does not always outperform the task-centric approach, we recommend that ML practitioners cautiously adapt the efficiency-centric ML system when the task accuracy is not absolutely important in their domain.
    
    \item We figured out the concatenated OOD detector significantly outperformed the prior task-centric paradigm. We examined the efficiency-centric paradigm's key strength is an overwhelming OOD detection performance compared to the task-centric approaches; thus, we highly recommend the ML practitioners utilize our efficiency-centric paradigm when OOD samples frequently exist.
    
    \item We experimentally validated the concatenated classifier is calibrated well to provide sufficient prediction results to the multi-labelable samples; therefore, we justified a efficiency-centric paradigm can reduce both the wasted computation resource  consumption and the inconvenience of end-users.
    
    \item We illustrated how the efficiency-centric paradigm is deployed in real-world live service at the large-scale car sharing platform. We expect the proposed deployment architecture will be a solid benchmark for candidate ML practitioners to apply in their domains.

\end{itemize}



\section{Preliminary}
\subsection{Background}
The core business of the car-sharing platform is enabling users to borrow a car with a smartphone application. When the users want to borrow a car, they can easily make a reservation on the smartphone application. Then, they go to the nearest parking station, open the car with a smartphone application, and start their trip. After using the car, the users park the car at the parking station where they borrowed it, and lock it on the smartphone to finish their use. During the aforementioned journey, SOCAR enforces the users to take pictures of the car and send them to the company through the application on particular events: before and after using the car, accident, car wash, charging washer fluids, and so on. To monitor the car's state without the human inspector's physical visit, human inspectors in SOCAR utilize these images to inspect the car's states. In SOCAR, there exist three primary car states that human inspectors are concerned with: \textit{Dirt}, \textit{Defect}, and \textit{Normal}. The \textit{Dirt} illustrates a state where the car's surface is dirty. The human inspectors are concerned with \textit{Dirt} state as dirty cars severely drop down the user experience and cause harsh claims. When the human inspectors recognize the car is in \textit{Dirt} state, they wash the car right away. The \textit{Defect} describes a car state where defects (i.e., scratch, dent) exist on the car's surface. The human inspectors are also concerned with \textit{Defect} as the damaged cars might cause the user's safety problems during the trip; thus, the inspectors send the car with any defects to the auto repair shop when they find it. The \textit{Normal} state implies that the car does not have any dirts or defects on its surface.

\subsection{Prior Image Recognition System} \label{sec:prior_image_tasks}

In this section, we aim to describe how the SOCAR has operated image recognition models before the introduction of \textit{efficiency-centric} paradigm. Along with the three car states mentioned above, there were two image classification systems: car dirt recognition and car defect recognition. The car dirt recognition model classifies between \textit{Normal} and \textit{Dirt} given a single image while the car defect-recognition model identifies between \textit{Normal} and \textit{Defect}. Following the \textit{task-centric} paradigm, we have acquired labeled datasets for each task, trained the models, established the OOD detectors, deployed them, and accumulated the prediction results in separated tables. We would like to highlight that every user-generated car image is accumulated in a single database. As the image recognition system performs a prediction daily, the image recognition system retrieves every image uploaded on a target day and provides them to the OOD detector and the trained classifier in sequential order.

\begin{figure}[htb!]
\centering
\begin{tabular} {c} \\
\includegraphics[width=0.4\textwidth, trim={0cm 0cm 0cm 0cm}, clip]{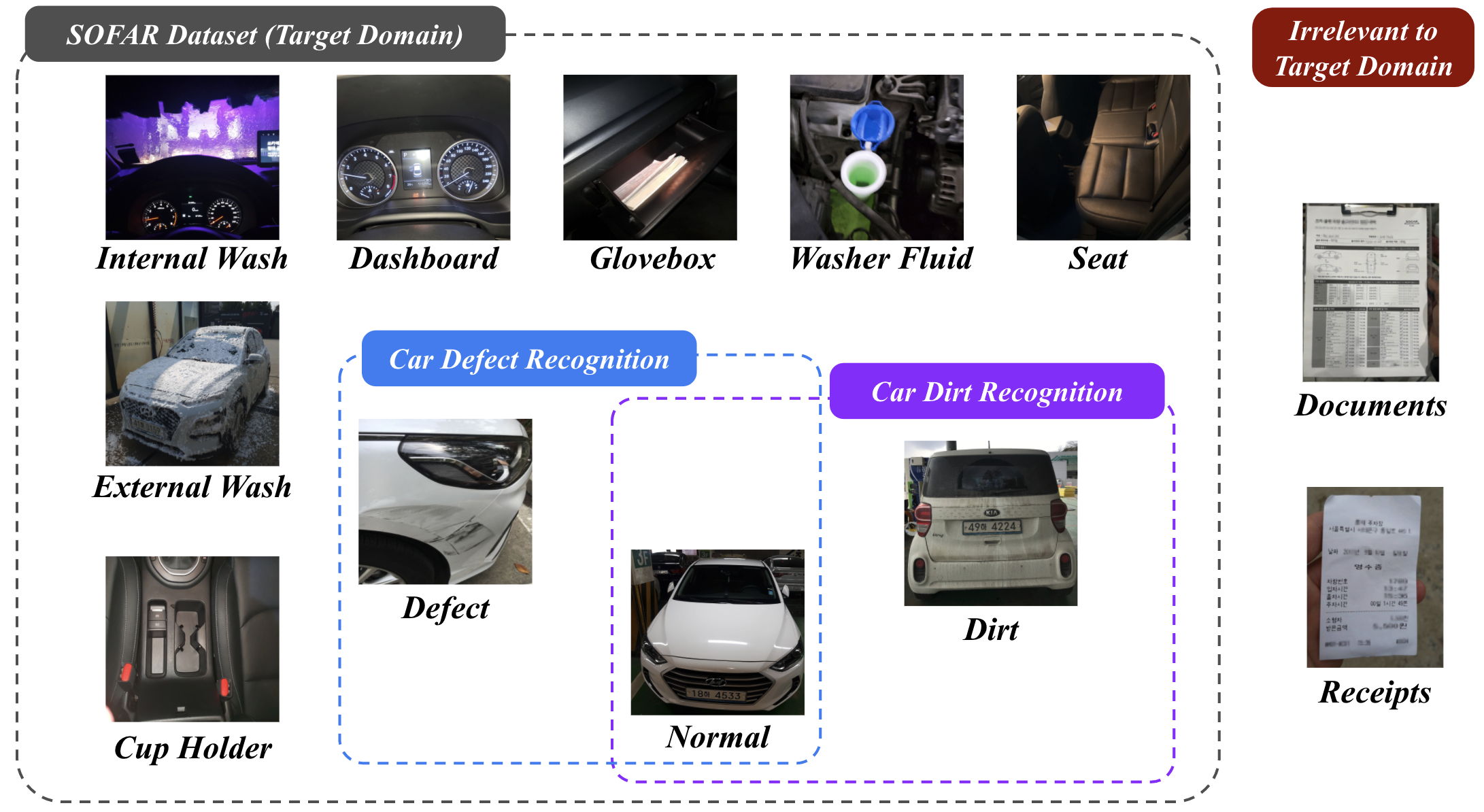} \\
\vspace{-0.5cm}
\end{tabular}
\caption{Image Recognition under SOFAR-Image Label space}
\label{fig:prior}
\end{figure}

In this image recognition system, we figured out several improvement avenues as described in section \ref{sec:intro}. For ease of understanding, we visualized the image recognition tasks in Figure \ref{fig:prior}. First, the ML practitioners have acquired two datasets for two tasks while they share the \textit{Normal} label simultaneously, as shown in Figure \ref{fig:prior}. Second, the prior image recognition system operates two OOD detectors for each task, although both tasks also share the same OOD samples that frequently exist in the database. Third, as both tasks retrieve inference samples from the same database, a single sample is provided into two classification systems while it does not have to be; thus, it causes a waste of computation resources. Lastly, as the prediction results are accumulated in separated tables, the human inspectors (end-users in Figure \ref{fig:approaches}) had to write two query scripts to understand the car's state. In a nutshell, we figured out the prior \textit{task-centric} ML systems bear several inefficiencies in the real-world environment. Motivated by this room for improvement, our study proposes a simple but effective approach to the ML system denoted as \textit{efficiency-centric} paradigm. 
While multi-task learning shares similar motivation with the proposed paradigm, it differs in the number of classification heads. Our efficiency-centric paradigm utilizes a classifier with a single classification head while multi-task learning embraces multiple heads. Our study presumes that concatenated head is much more advantageous in representation learning than utilizing multiple heads. A unified head under cross-entropy loss lets the model choose the most probable label; thus, it can enforce the model to be more focused on discriminative characteristics among labels.

\section{Efficiency-centric ML System}
\subsection{Description}
In this section, we elaborate on how we established \textit{efficiency-centric} ML system in the car sharing platform domain. The development of \textit{efficiency-centric} ML system consists of three steps: dataset acquisition, model training, and establishing an OOD detector. The detailed descriptions of each step are illustrated in the following sections.

\subsubsection{Dataset Acquisition}
\begin{figure*}[htb!]
\centering
\begin{tabular} {c} \\
\includegraphics[width=0.8\textwidth, trim={0cm 0cm 0cm 0cm}, clip]{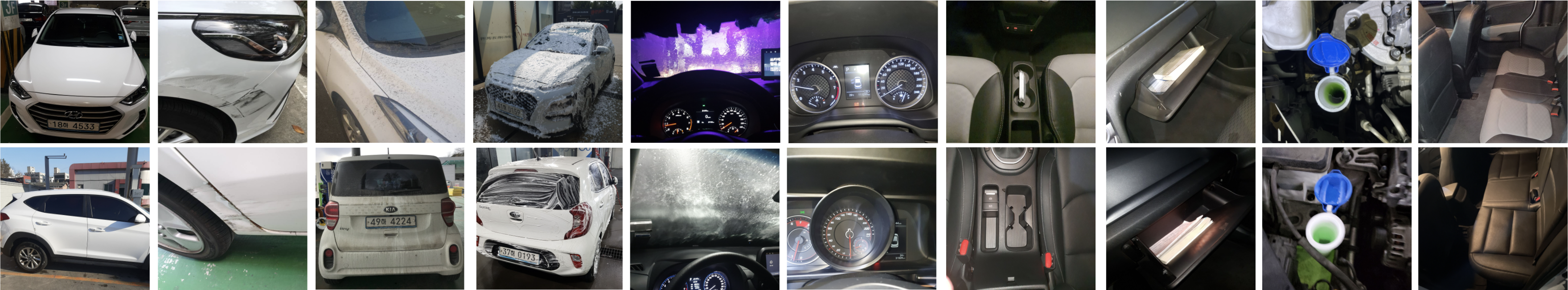} \\
\vspace{-0.6cm}
\end{tabular}
\caption{Samples from the SOFAR-Image. Each column illustrates an individual label of the dataset. From the the very left, each column represents \textit{Normal}, \textit{Defect}, \textit{Dirt}, \textit{Bubble Wash}, \textit{Cars inside of the Washing Machine}, \textit{Dashboard}, \textit{Cup Holder}, \textit{Glovebox}, \textit{Washer Fluid}, \textit{Seat}.}
\label{fig:sofar}
\end{figure*}

First and foremost, we acquired a domain-specific benchmark dataset by concatenating every \textit{task-centric} dataset. Practically, we investigated every acquired dataset at SOCAR, clarified the labeling scheme, and concatenated the datasets into a single one. For example, given car dirt recognition dataset (\textit{Normal} v. \textit{Dirt}) and car defect recognition dataset (\textit{Normal} v. \textit{Defect}) for the tasks described in section \ref{sec:prior_image_tasks}, we concatenated them as a single benchmark dataset (\textit{Normal} v. \textit{Dirt} v. \textit{Defect}). Not only these three labels, throughout multiple interviews with human inspectors (the ML system's end-users), we discovered eight more labels that frequently exist in a database: \textit{Bubble Wash}, \textit{Cars inside the Car Wash Machine}, \textit{Dashboard}, \textit{Cup Holder}, \textit{Glovebox}, \textit{Washer Liquid}, and \textit{Seats}. Therefore, we established a domain-specific benchmark dataset consisting of 10 labels, and we denoted this benchmark dataset as \textbf{SOcar dataset For Advanced Research (SOFAR)-Image}. Note that samples at each label are visualized in the Figure \ref{fig:sofar}. Additionally, the dataset can be accessed in \url{https://socar-kp.github.io/sofar_image_dataset/} with detailed descriptions. 

\subsubsection{Model Training}
Given a SOFAR-Image, we trained an 10-class classifier following a conventional image recognition system. As the prior image recognition systems discovered that deep neural networks architecture in fine-grained recognition \cite{wei2019deep,wang2018learning,zhao2017survey} were effective in our domain, we employed Progressive Multi-Granularity network (PMG) \cite{du2020finegrained} as a model. We set the learning objective as minimizing conventional cross-entropy loss. As implementation details, we set the batch size of 16, learning rate of 2e-3, weight decay with the parameter of 5e-4. We empirically applied cosine annealing \cite{loshchilov2017sgdr} as a learning rate scheduler to optimize the parameters effectively. Note that we publicized our codes in \url{https://github.com/socar-tigger/kdd_2022_submission} for the reproducibility of our study. 

\subsubsection{Establishing the OOD Detector}
Lastly, we established an OOD detector following the conventional approach \cite{hendrycks2018baseline} in the public domain datasets at the OOD detection studies. We extracted the maximum logit value at the layer right before the softmax layer, denoting this as confidence. Given a validation sample, we regard the sample where its confidence resides lower than a pre-set threshold as an OOD. As the pre-set threshold influences the OOD detection performance, we set the evaluation metrics of the OOD detector as an Area Under the ROC curve (AUROC) to check overall OOD detection performance over various threshold levels properly.

\subsection{Research Questions}
We postulated several research questions to examine whether the \textit{efficiency-centric} ML system can substitute the prior \textit{task-centric} ML system. We set the \textit{efficiency-centric} ML system should fulfill three technical requirements to substitute the prior \textit{task-centric} ML system. These research questions are described as follows.

\begin{itemize}
    \item \textbf{RQ 1. Does the \textit{efficiency-centric} ML system accomplish comparable task accuracy to the \textit{task-centric } approach?}: Our study aimed to examine whether the \textit{efficiency-centric} approach can achieve competitive task accuracy with the \textit{task-centric} approach. 

    \item \textbf{RQ 2. Does the \textit{efficiency-centric} ML system perform  OOD detection better rather than \textit{task-centric} approach?}: We set experiments to examine whether the proposed \textit{efficiency-centric} ML system accomplishes better OOD detection compared to the \textit{task-centric} paradigm. 

    \item \textbf{RQ 3. Does the Top-\textit{n} predictions \textit{efficiency-centric} ML system yield correct labels on a multi-labelable sample?}: Lastly, RQ 3. justifies the substitutability of our approach on the concatenation of prediction result tables. 
    
    We presume a single, concatenated prediction result table can substitute multiple \textit{task-centric} tables if the efficiency-centric classifier is calibrated well enough to predict multi labels on the multi-labelable samples. 
    
\end{itemize}

\subsection{Baseline for Comparative Study}

\subsubsection{Task-Centric ML System}
The task-centric ML system is a conventional approach to the image recognition task. Given a labeled dataset regarding the task, (i.e., \textit{Normal} and \textit{Defect} at SOFAR-Image to the car defect recognition), the task-centric paradigm trains the classifier under the supervised regime. To establish an OOD detector, we followed the same OOD detection method \cite{hendrycks2018baseline} with the \textit{efficiency-centric} ML system for a proper comparative study.

\subsubsection{Multi-Task Learner}
The multi-task learner is an intuitive way to handle multiple tasks and be computationally efficient. We employed this multi-task learner as a baseline of our study as it jointly solves several tasks given a single backbone neural network; thus, it can be a candidate alternative to the \textit{task-centric} ML system. We design a conventional multi-heads classifier with a CNN backbone and a few heads for each task for comparison with the efficiency-centric ML system. As a classifier, we use four heads; three heads are for binary classification(\textit{Defect} v. \textit{Normal}, \textit{Dirt} vs \textit{Normal}, \textit{Bubble Wash} vs \textit{Normal}) and one head for multi-class classification on the other labels of SOFAR-Image. For training, binary cross-entropy loss with sigmoid function is used for the binary classification heads, and cross-entropy loss with softmax function is used for a multi-class classification head. The training loss is a summation of losses that occurred from each head. In the evaluation phase, we only use the prediction of the target task head. For example, when we desire to investigate defect recognition performance of the multi-task learner, other predictions except the defect task head are ignored. 

Unfortunately, the OOD detection method \cite{hendrycks2018baseline} utilized in both task-centric and efficiency-centric ML systems cannot be directly applied to the multi-task leaner due to its network architecture. Given an \textit{Normal} sample to the trained multi-task learner, the output logits of the binary heads would be nearly zero (implying an OOD) even though it distributes within the training set. To resolve this limit, we utilized the Kullback-Leibler (KL) divergence between a uniform distribution and the concatenated logit of the multi-task learner. We presume an OOD sample should not be activated in every head of the multi-task learner. We expect a sample with large KL divergence implies that the model understands particular characteristics of the image; thus, the sample is adjacent to the training set distribution. For the sample with small KL divergence, vice versa. For example, given an OOD sample to the multi-task learner, the distribution of predicted logit shall be similar to the uniform distribution as follows. A logit at each binary head should be nearly 0.5, and the one from the multi-class head shall be 1 / number of classes on the multi-class head.

\section{Does the efficiency-centric ML system solve the Target Task Well?} \label{sec:task}
\subsection{Setup}
In pursuit of discovering an answer to the \textbf{RQ 1.}, we aim to examine whether the \textit{efficiency-centric} ML system can accomplish promising task performance compared to the \textit{task-centric} approach. As there exist more labels in \textit{engineerning-centric} approach, we presumed the proposed ML system is exposed to the higher risk of making a faulty prediction. Suppose that we solve a car defect recognition task (\textit{Normal} v. \textit{Defect}). Under the \textit{task-centric} paradigm, the model would learn a representation power biased to the two labels only. It implies that the model learns discriminative characteristics of \textit{Defect} label relative to the \textit{Normal}, and vice versa; thus, the model shall scrutinize only two label spaces at the end of the training. On the other hand, under the \textit{efficiency-centric} paradigm, the model would acquire the representation power, which can classify more labels (in the SOFAR-Image dataset). This widened label space might contribute to more qualified representation power for the target class (\textit{Defect}) or disqualify it by adding unnecessary knowledge to the task. To clarify it, we designed experiments to examine how the \textit{efficiency-centric} ML system accomplishes task performances at two target image recognition tasks: car defect recognition and car dirt recognition.

We trained the proposed efficiency-centric classifier (11-class classifier) and two benchmark classifiers: task-centric classifier and multi-task classifier. We established two evaluation suites throughout the experiments: SOFAR-Image-Test and several external validation datasets. The SOFAR-Image-Test implies a test set at the SOFAR-Image, which shares the same distribution with the training set; thus, we presumed the SOFAR-Image-Test does not bear much dataset shift \cite{fang2020rethinking} from the training set. On the other hand, to validate our efficiency-centric ML system's robustness, we also acquired external validation datasets that have a particular dataset shift from the training set. The SOFAR-Image includes samples of white cars the users took on a sunny day, which implies less noisy patterns from the rain or snow. In contrast, we established external validation datasets consisting of car images taken in various weather conditions. We empirically presume car images affected by various weather conditions bear noisy patterns (i.e., snow or raindrops on the car's surface); thus, these samples are adequate to evaluate the robustness of our approach. We established four external validation sets for car defect recognition (\textit{Ext-Snow}, \textit{Ext-After Snow}, \textit{Ext-Rain}, \textit{Ext-After Rain}) and two external validation set for car dirt recognition (\textit{Ext-H1}, \textit{Ext-H2}) for car dirt recognition. Note that H1 implies the time horizon from January to June, and the H2 describes the other months. We set these external validation sets following the guide provided by the human inspectors in the car sharing platform. They noted that the defect detection should be robust in various weather conditions while the dirt detection had better be robust in different time horizons. We measured the target Accuracy and F1-score at each task as an evaluation metric. We recorded the mean and the standard deviations of each evaluation metric on three trials. The experiment results are shown in Table \ref{table:result_acc}. and Table \ref{table:result_f1}.

\begin{table*}[htb!]
\centering
\caption{\textit{Accuracy} on various image recognition performances under several evaluation suites consisting of the SOFAR-Image-Test and series of external validation sets. While our \textit{efficiency-centric} classifier accomplished superior performance in car defect recognition, it achieved competitive performance rather than baseline approaches.}
\resizebox{0.95\textwidth}{!}{
\begin{tabular}{|c|c|c|c|c|c|c|c|c|}

\hline
\multirow{3}{*}{\textbf{Method}} & \multicolumn{8}{|c|}{\textbf{Image Recognition Tasks}} \\
\cline{2-9}
 & \multicolumn{5}{|c|}{\textbf{Car Defect Recognition}} & \multicolumn{3}{|c|}{\textbf{Car Dirt Recognition}} \\
\cline{2-9}
 & \textbf{SOFAR-Test} & \textbf{Ext-Snow} & \textbf{Ext-After Snow} & \textbf{Ext-Rain} & \textbf{Ext-After Rain} & \textbf{SOFAR-Test} & \textbf{Ext-H1} & \textbf{Ext-H2} \\
\hline

\textbf{efficiency-centric (OURS)}&\textbf{0.930 $\pm$ 0.017} & \textbf{0.691 $\pm$ 0.022} & \textbf{0.728 $\pm$ 0.016} & \textbf{0.739 $\pm$ 0.021} & \textbf{0.787 $\pm$ 0.017}  & $0.937 \pm 0.010$ & $0.907 \pm 0.007$ & $0.829 \pm 0.007$ \\
\textbf{Multi-Task Learning}& $0.918 \pm 0.011$ & $0.567 \pm 0.012$ & $0.655 \pm 0.015$ & $0.712 \pm 0.023$ & $0.766 \pm 0.008$  & \textbf{0.957 $\pm$ 0.003} & $0.898 \pm 0.010$ & \textbf{0.858 $\pm$ 0.003} \\
\textbf{Task-Centric-PMG}& $0.917 \pm 0.023$ & $0.622 \pm 0.072$ & $0.671 \pm 0.064$ & $0.692 \pm 0.059$ & $0.724 \pm 0.058$ & $0.962 \pm 0.002$ & \textbf{0.909 $\pm$ 0.001} & $0.852 \pm 0.002$ \\
\hline

\end{tabular}
}
\label{table:result_acc}
\end{table*}

\begin{table*}[htb!]
\centering
\caption{\textit{F1 score} on various image recognition performances under several evaluation suites consisting of the SOFAR-Image-Test and series of external validation sets. While our \textit{efficiency-centric} classifier accomplished superior performance in car defect recognition, it achieved competitive performance rather than baseline approaches.}
\resizebox{0.95\textwidth}{!}{
\begin{tabular}{|c|c|c|c|c|c|c|c|c|}

\hline
\multirow{3}{*}{\textbf{Method}} & \multicolumn{8}{|c|}{\textbf{Image Recognition Tasks}} \\
\cline{2-9}
 & \multicolumn{5}{|c|}{\textbf{Car Defect Recognition}} & \multicolumn{3}{|c|}{\textbf{Car Dirt Recognition}} \\
\cline{2-9}
 & \textbf{SOFAR-Test} & \textbf{Ext-Snow} & \textbf{Ext-After Snow} & \textbf{Ext-Rain} & \textbf{Ext-After Rain} & \textbf{SOFAR-Test} & \textbf{Ext-H1} & \textbf{Ext-H2} \\
\hline

\textbf{efficiency-centric  (OURS)}& \textbf{0.930 $\pm$ 0.017} & \textbf{0.509 $\pm$ 0.014} & \textbf{0.599 $\pm$ 0.013} & \textbf{0.614 $\pm$ 0.012} & \textbf{0.692 $\pm$ 0.016}  & $0.934 \pm 0.011$ & $0.817 \pm 0.016$ & $0.781 \pm 0.011$ \\
\textbf{Multi-Task Learning}& $0.918 \pm 0.011$ & $0.442 \pm 0.007$ & $0.548 \pm 0.011$ & $0.612 \pm 0.016$ & $0.675 \pm 0.007$  & $0.956 \pm 0.003$ & \textbf{0.843 $\pm$ 0.011} & \textbf{0.843 $\pm$ 0.001} \\
\textbf{Task-Centric-PMG}& $0.916 \pm 0.023$  & $0.474 \pm 0.037$ & $0.557 \pm 0.040$ & $0.593 \pm 0.040$  & $0.640 \pm 0.045$ & \textbf{0.962 $\pm$ 0.003} & $0.835 \pm 0.004$ & $0.819 \pm 0.004$ \\
\hline

\end{tabular}
}
\label{table:result_f1}
\end{table*}

\subsection{Experiment Results}
Along with the experiment results, we discovered that the proposed \textit{efficiency-centric} paradigm accomplishing competitive task performance in two image recognition tasks, but there still exists room for improvement in the car dirt recognition task. We analyzed these results occurred because the \textit{efficiency-centric} classifier acquired more general, knowledgeable representation power rather than other baselines. A prominent characteristic of \textit{task-centric} classifier is that it can create a representation power particularly biased to the given classes. Referring to car defect recognition task, the \textit{task-centric} classifier learns the knowledge focused only at \textit{Normal} and \textit{Defect}. The multi-task learner also learns knowledge biased to the given classes as the last head performs a binary classification (\textit{Defect} v. \textit{No  Defect}). Conversely, the proposed \textit{efficiency-centric} classifier acquires the representation power less-biased to particular classes. As it scrutinizes wider representation space from the SOFAR-Image dataset (including 11 labels), we analyzed our \textit{efficiency-centric} classifier learned more generalizable representation power compared to the baselines. This comparatively general representation power was more advantageous in solving a car defect recognition but less effective in the car dirt recognition. To paraphrase it, more general representation power can fulfill sufficient task accuracy at each task, but sometimes the biased representation power could be much more advantageous. Note that scrutinizing this relationship between the representation power's characteristics and task accuracy is an improvement avenue of our study. Nevertheless, we evaluate the proposed \textit{efficiency-centric} classifier achieved competitive task performances compared to other benchmarks. Practically, human inspectors at SOCAR evaluated our \textit{efficiency-centric} classifier's performance on SOFAR-Test as sufficient enough to be applied in the real world; thus, we decided to integrate it.

\subsection{Recommendations} \label{sec:task_recommendation}
As a lesson learned from the experiment, we highly recommend candidate ML practitioners to establish \textit{efficiency-centric} classifier when their domain does not require perfect task performance. The car sharing platform can comparatively allow the incorrect prediction of the machine learning model as it does not cause excessively fatal damage. Moreover, damage from the faulty predictions is mitigated as the human inspectors check the prediction results before the business operation. Thus, we expect candidate ML practitioners in the domain similar to our situation can utilize the proposed \textit{efficiency-centric} paradigm as it acquires competitive performance to the prior \textit{task-centric} paradigm. On the other hand, we recommend ML practitioners in accuracy-prioritized domains (i.e., medical applications) should be cautious in utilizing the proposed paradigm as the degraded task performance might cause serious damage to the end-users. 

\section{Does the efficiency-centric Classifier solve the OOD Detection Better?}
\subsection{Setup}
As an answer to \textbf{RQ 2.}, our study validated whether the proposed \textit{efficiency-centric} ML system achieves better OOD detection performance rather than baselines. We expect the \textit{efficiency-centric} ML system would solve the OOD detection better as its representation is less biased to particular classes; thus, it can understand the characteristics of samples irrelevant to the target task and reject them. To examine the effectiveness of the proposed \textit{efficiency-centric} ML system, we postulated two OOD detection tasks: One-class recognition and irrelevant sample rejection.

\begin{figure}[htb!]
\centering
\begin{tabular} {c c} \\
\includegraphics[width=0.13\textwidth, trim={0cm 0cm 0cm 0cm}, clip]{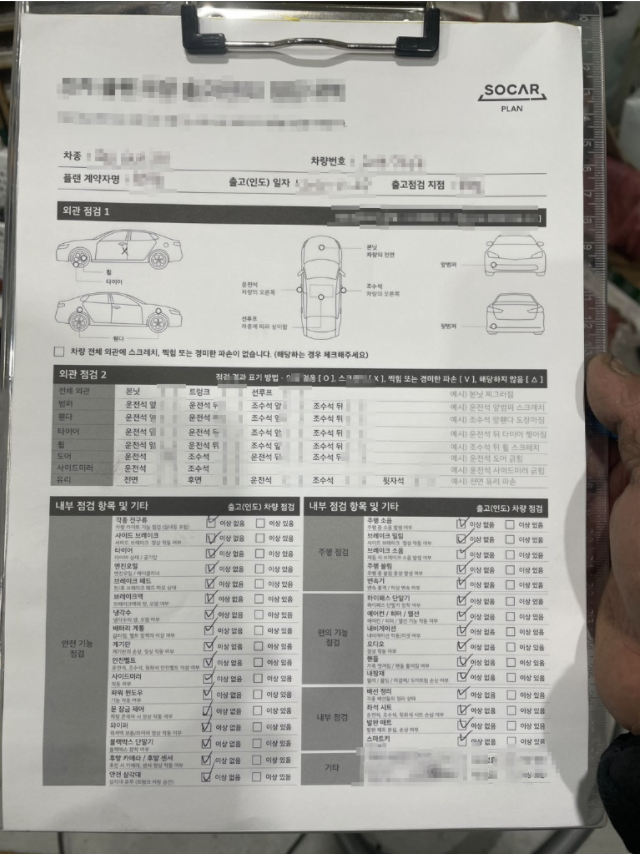} &
\includegraphics[width=0.13\textwidth, trim={0cm 0cm 0cm 0cm}, clip]{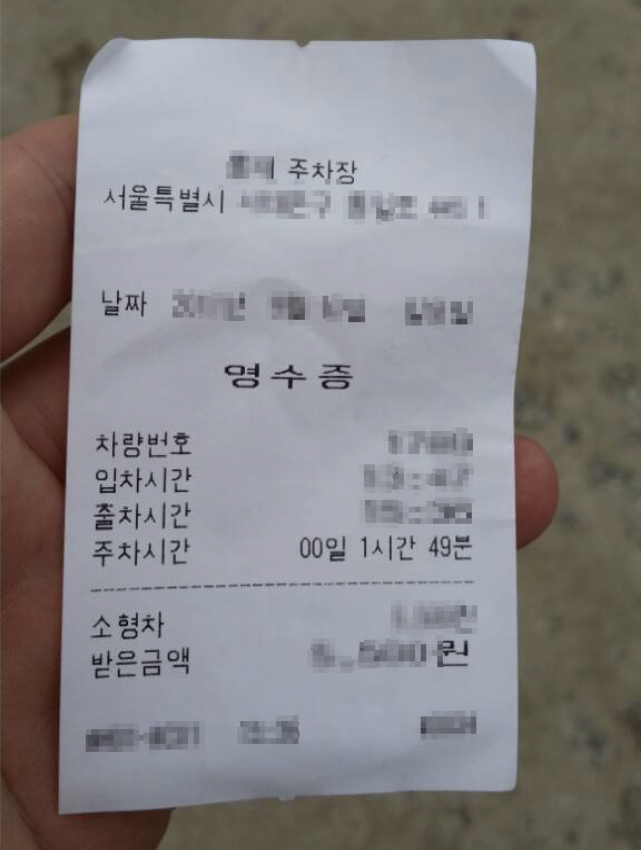} \\
\vspace{-0.1cm}
{\small (a) Document } & {\small (b) Receipt } \\
\end{tabular}
\caption{Samples in OOD labels (\textit{Document} and \textit{Receipts}).
}
\label{fig:real_ood_samples}
\end{figure}

\subsubsection{One Class Recognition}
First, the one-class recognition aims to validate whether the efficiency-centric ML system effectively identifies the target label among the labels in the SOFAR-Image-Test dataset. Our study designed a one class recognition experiment upon two target labels: \textit{Defect} and \textit{Dirt}. Regarding the \textit{Defect} label, given an inference dataset (SOFAR-Image-Test), we set the model to discriminate \textit{Defect} samples from the other labels; thus, we let the model solve \textit{one v. all} recognition task. On the \textit{Dirt} label, the model shall identify \textit{Dirt} samples from the other labels. We interpret the higher one-class recognition performance implies the better OOD detection performance as it can effectively discriminate the target label from the other labels. As an evaluation metric for one-class recognition, we use the AUROC along with various OOD detection thresholds. The experiment result is illustrated in Table \ref{table:one_class_recognition}.

\subsubsection{Irrelevant Sample Rejection} \label{sec:irrelevant}

Furthermore, we further examined whether the \textit{efficiency-centric} ML system effectively rejects samples irrelevant to the domain. In the real world, there frequently exists samples that are not related to the target domain. While we deal with labels regarding the car's state based on the SOFAR-Image dataset, the human inspectors frequently discovered images not relevant to the car's state, as shown in Figure \ref{fig:real_ood_samples} If these samples are not rejected in the OOD detector, the model will yield a faulty prediction result on the irrelevant sample, which deteriorates the end-user's business operations. Therefore, our study retrieved about 100 samples at two irrelevant labels (\textit{Receipts} and \textit{Documents}) from the database and checked whether the proposed \textit{efficiency-centric} ML system precisely rejects them. We also employed the AUROC for evaluation.
The experiment results are shown in Table \ref{table:irrlevant}.

\subsection{Experiment Results} \label{sec:ood}
Along with the OOD detection performance in Table \ref{table:one_class_recognition} and Table \ref{table:irrlevant}, we resulted in the proposed \textit{efficiency-centric} ML system is surprisingly outperformed the \textit{task-centric} approach and multi-task learner. We analyzed that this significant performance derives from the wider representation power of the proposed ML system compared to the other benchmarks. As we noted in section \ref{sec:task_recommendation}, one primary characteristic of \textit{task-centric} paradigm is biased representation power on the given classes. While this characteristic was advantageous in accomplishing higher task accuracy, however, we analyze this biased representation becomes a disadvantage in OOD detection. Referring to numerous studies on OOD detection \cite{yang2021generalized,hendrycks2018baseline}, a key element of an effective OOD detector is a qualified representation power that can understand samples in various labels, even if they exist outside of the domain. As the \textit{task-centric} ML system aims to establish a biased representation to the target class only, we analyze the representation power cannot understand OOD samples properly. Not only the \textit{task-centric} approach, but the multi-task learning approach also lacks the quality of representation power as its biasedness exists in multiple heads that solve each task. On the other hand, our \textit{efficiency-centric} ML system lets the model explore wider representation space based on the concatenated dataset (SOFAR-Image); thus, the model can acquire more qualified representation power which understands more general characteristics of various samples. Consequentially, we resulted in the proposed \textit{efficiency-centric} ML system being more robust to OOD samples in a real world setting. In a practical manner, we analyzed this superficial OOD detection performance is a key advantage of our \textit{efficiency-centric} ML system.

\begin{table}[htb!]
\centering
\caption{One-Class Recognition Performance}
\resizebox{0.38\textwidth}{!}{
\begin{tabular}{|c|c|c|}

\hline
\multirow{2}{*}{Method} & \multicolumn{2}{|c|}{\textbf{AUROC}} \\
\cline{2-3}
& \textbf{Dirt} & \textbf{Defect} \\
\hline

\textbf{Efficiency-centric (OURS)} & \textbf{0.998} & \textbf{0.995}  \\
\textbf{Multi-Task Learning} & 0.938 & 0.873  \\
\textbf{Task-Centric} & 0.993 & 0.971  \\
\hline

\end{tabular}
}
\label{table:one_class_recognition}
\end{table}

\begin{table}[htb!]
\centering
\caption{Irrelevant Sample Rejection Performance}
\resizebox{0.38\textwidth}{!}{
\begin{tabular}{|c|c|c|}

\hline
\multirow{2}{*}{Method} & \multicolumn{2}{|c|}{\textbf{AUROC}} \\
\cline{2-3}
& \textbf{Receipts} & \textbf{Documents} \\
\hline

\textbf{Efficiency-centric (OURS)} & \textbf{0.958} & \textbf{0.971}  \\
\textbf{Multi-Task Learning} & 0.604 & 0.577 \\
\textbf{Task-Centric (Dirt)} & 0.884 & 0.853  \\
\textbf{Task-Centric (Defect)} & 0.841 & 0.817  \\
\hline

\end{tabular}
}
\label{table:irrlevant}
\end{table}





\subsection{Recommendations}
Following the experiment results, we recommend the candidate ML practitioners to employ our \textit{efficiency-centric} ML system when their domain prioritizes an OOD detection performance. Along with our recommendations proposed in Section \ref{sec:task}, the \textit{task-centric} paradigm's biased representation power is a double-edged sword. While the biased representation power of the \textit{task-centric} classifier accomplishes higher task accuracy, it also has less robustness toward the OOD samples. On the other hand, the proposed \textit{efficiency-centric} paradigm achieves competitive performance to the \textit{task-centric} approach, but it guarantees significant robustness against the OOD samples. We would like to highlight a trade-off between \textit{task-centric} paradigm and \textit{efficiency-centric} paradigm to the candidate ML practitioners. If their domain does not include many OOD samples (i.e., the data source is well-controlled not to create OOD samples), they might adapt \textit{task-centric} ML system to enjoy the precise task accuracy. Conversely, suppose their domain bears many OOD samples similar to our car sharing platform (i.e., the data source is hardly controlled or user-generated samples on a large scale). In that case, we highly recommend the to adapt the proposed efficiency-centric ML system with competitive task accuracy and superficial OOD detection performance.

\section{Does the efficiency-centric Classifier Calibrate Well?}

\subsection{Setup}
\begin{figure}[htb!]
\centering
\begin{tabular} {c} \\
\includegraphics[width=0.4\textwidth, trim={0cm 0cm 0cm 0cm}, clip]{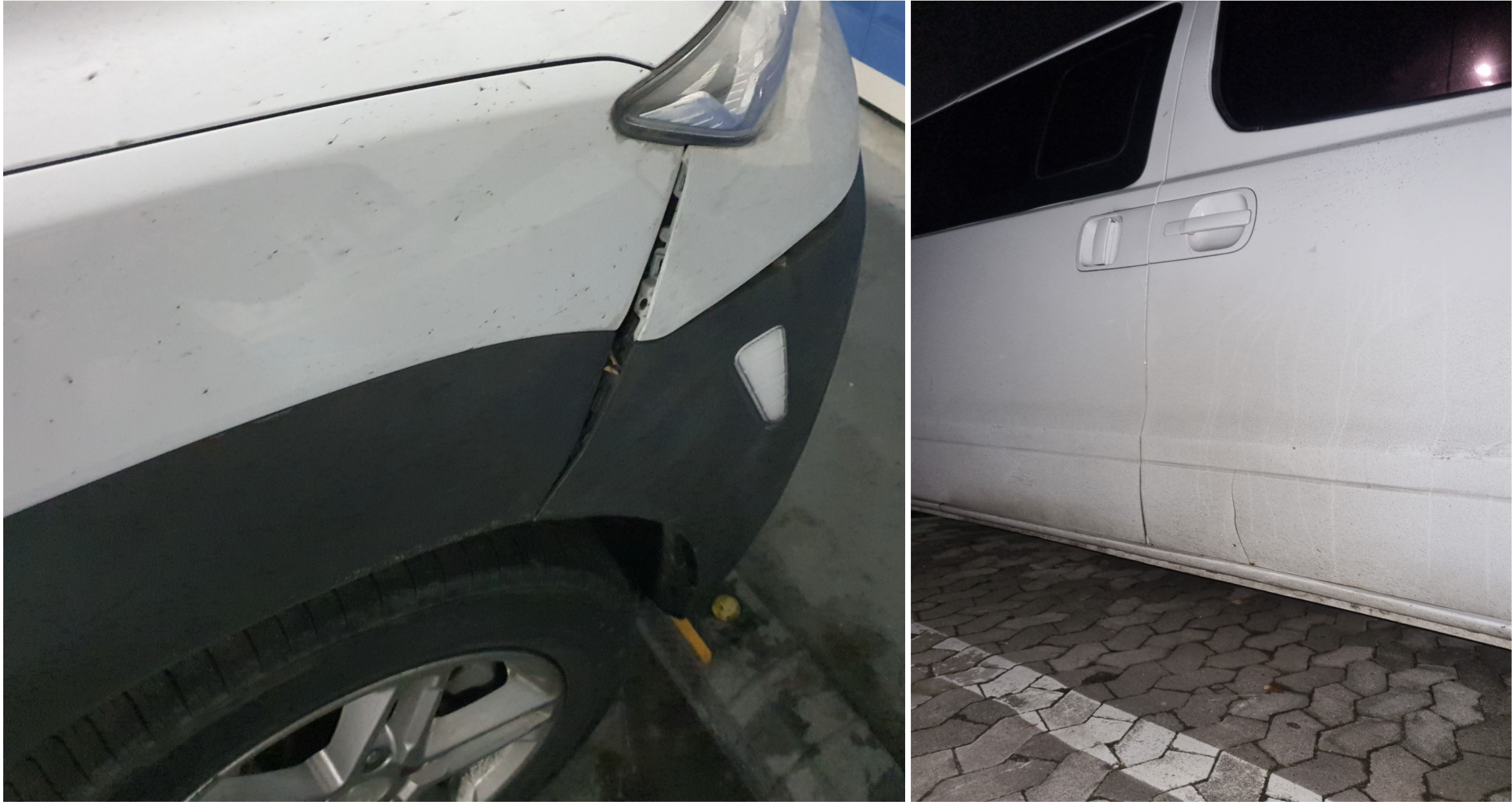} \\
\vspace{-0.7cm}
\end{tabular}
\caption{Example images of the multi-labelable samples
}
\label{fig:multi}
\end{figure}

Last but not least, we aimed to validate whether the \textit{efficiency-centric} ML system can yield sufficient prediction results on a multi-labelable sample. The multi-labelable sample is a sample that has attributes of two labels simultaneously. In a car sharing platform, as shown in Figure \ref{fig:multi}., a multi-labelable image frequently exists that has both dirt and defect on its surface. Suppose these multi-labelable samples exist in the inference dataset. Under the \textit{task-centric} ML system, a single ML system can only identify one state (label) of the given multi-labelable sample. Referring to car defect recognition system and car dirt recognition system, the defect recognition system can only identify \textit{Dirt} while the dirt recognition system can only recognize \textit{Defect}. As human inspectors (end-users) desire to acquire both prediction results (\textit{Defect} and \textit{Dirt}), the prior \textit{task-centric} paradigm repetitively provide the inference dataset into two ML systems (defect recognition and dirt recognition). Then, prediction results at each task are accumulated in separated prediction result tables, and the end-user has to query every table to utilize prediction results.

\begin{figure}[htb!]
\centering
\begin{tabular} {c} \\
\includegraphics[width=0.3\textwidth, trim={0cm 0cm 0cm 0cm}, clip]{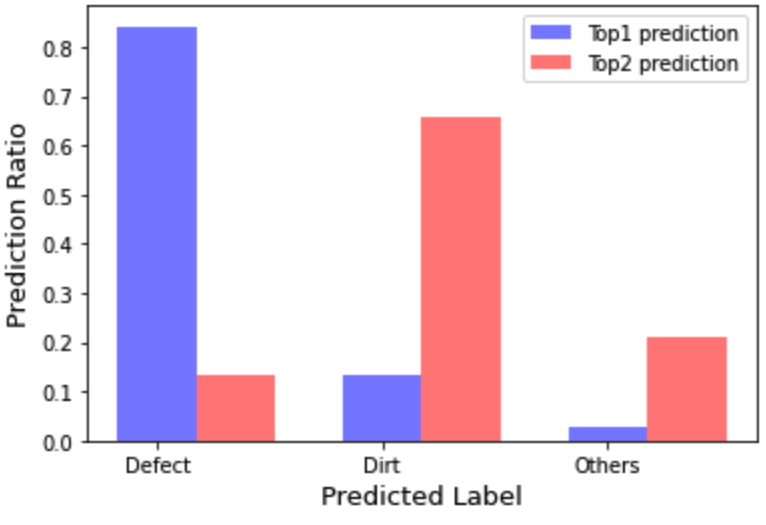} \\
\vspace{-0.9cm}
\end{tabular}
\caption{Top-2 predictions of our \textit{efficiency-centric} classifier
}
\label{fig:multi_result}

\end{figure}

We figure out several drawbacks upon the aforementioned procedures dealing with multi-labelable samples. First, one sample has to be inferenced twice to provide multi-labels, and it creates a particular amount of resource consumption. Second, the end-user should query multiple tables to retrieve prediction results (\textit{Defect} and \textit{Dirt}) on a single sample, which causes inconvenience. Therefore, we aim to examine whether our \textit{efficiency-centric} ML system can provide adequate prediction results on a multi-labelable sample only with a single inference. As the proposed classifier learned the characteristics of both \textit{Defect} and \textit{Dirt}, we expected the Top-2 prediction results of the \textit{efficiency-centric} classifier would include \textit{Defect} and \textit{Dirt} label. If our expectation is valid, we can perform inference for a single time on the inference dataset and accumulate Top-2 prediction results in a single table; this would reduce resource consumption and provide convenience to the end-users. Therfore, we acquired 100 multi-labelable samples of \textit{Defect} and \textit{Dirt} and examined whether the \textit{efficiency-centric} classifier yield these two labels in Top-2 prediction results. The experiment result is illustrated in Figure \ref{fig:multi_result}.

\begin{figure}[htb!]
\centering
\begin{tabular} {c} \\
\includegraphics[width=0.3\textwidth, trim={0cm 0cm 0cm 0cm}, clip]{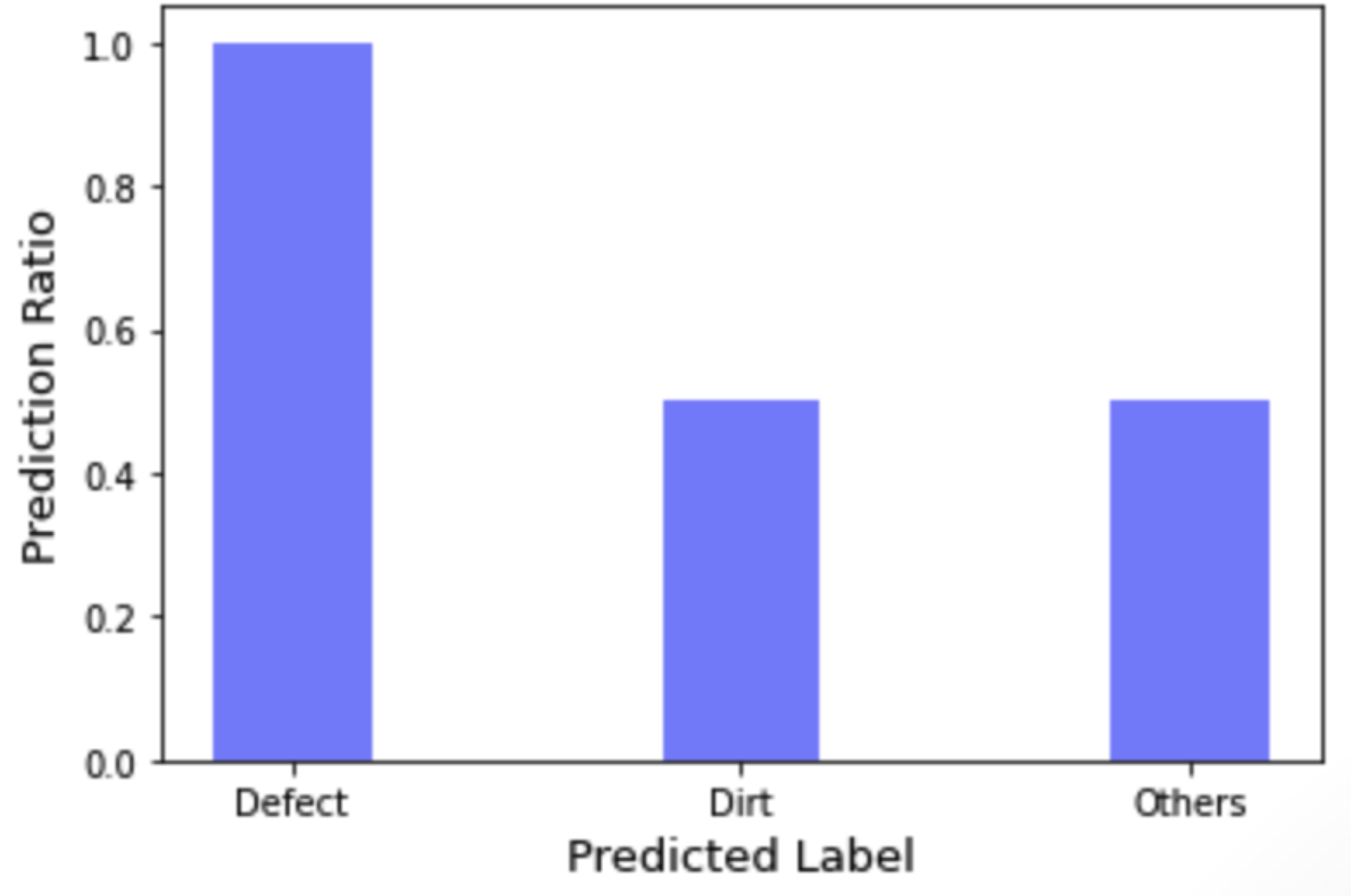} \\
\vspace{-0.9cm}
\end{tabular}
\caption{Prediction results of the \textit{multi-task learner}
}
\label{fig:multi_result_benchmark}
\end{figure}

\subsection{Experiment Results}
Based on the experiment result shown in Figure \ref{fig:multi_result} and Figure \ref{fig:multi_result_benchmark}., we resulted in our \textit{efficiency-centric} classifier can provide satisfactory prediction results on a multi-labelable sample. Given multi-labelable samples of \textit{Defect} and \textit{Dirt}, majority of Top-2 prediction results include these two labels. We expect that this result derived from the widened representation power of the proposed \textit{efficiency-centric} classifier. As the classifier learns each label's characteristics from the SOFAR-Image, within the top-2 predictions, it properly predicted both labels (\textit{Defect} and \textit{Dirt}) if a given image has attributes of two labels. Note that the gap between prediction results of \textit{Defect} and \textit{Dirt} is not particularly large on the efficiency-centric classifier, which implies the representation power is calibrated well. On the other hand, we figured out that multi-task learned succeeded in predicting \textit{Defect} on the multi-labelable samples while it could not effectively recognize \textit{Dirt}. Moreover, the gap between predictions on \textit{Defect} and \textit{Dirt} is comparatively larger than the one in efficiency-centric classifier. We expect this different calibration performance between our efficiency-centric classifier and the multi-task learner derived from the representation power. As we have described in the prior section \ref{sec:ood}, the multi-task learner is biased to particular classes compared to the efficiency-centric classifier. Upon the result shown in Figure \ref{fig:multi_result_benchmark}, we expect the representation power of multi-task learner is much biased to the characteristics of \textit{Defect} rather than \textit{Defect}. Thus, it creates a significant understanding of the \textit{Defect} label, but comparatively worsens the knowledge on the \textit{Dirt} label. Conversely, our efficiency-centric classifier yields competitive prediction results on both labels as long as its representation power is less biased to the particular label. In a nutshell, we validated that the proposed \textit{efficiency-centric} classifier can provide adequate multiple prediction results on a multi-label label sample; thus, it can substitute the \textit{task-centric} ML system to reduce the waste of computation resource and enhance the convenience of end-users.

\begin{figure*}[htb!]
\centering
\begin{tabular} {c} \\
\includegraphics[width=0.7\textwidth, trim={0cm 0cm 0cm 0cm}, clip]{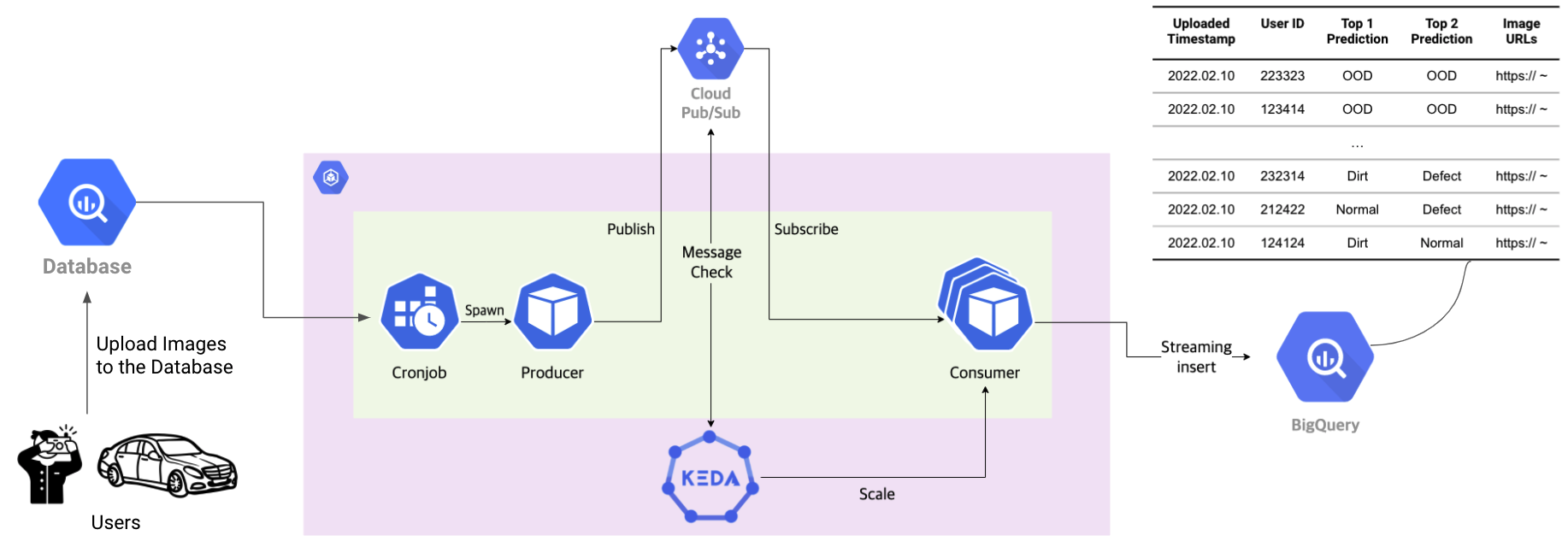} \\
\vspace{-0.5cm}
\end{tabular}
\caption{An architecture of the deployed efficiency-centric ML system in a cloud environment.}
\label{fig:deploy_final}
\end{figure*}

\section{Real-world deployment}
Throughout the following sections, we discovered a trade-off between task accuracy and engineering efficiency regarding prior \textit{task-centric} ML system and the proposed \textit{efficiency-centric} ML system. While the \textit {task-centric} ML system provides a precise task performance, it lacks engineering efficiency in two aspects: 1) The number of ML pipeline increases linear to the number of tasks, 2) As the \textit{task-centric} ML system is not robust to OOD samples, the data pipeline should be well-controlled. Conversely, our \textit{efficiency-centric} ML system yields an engineering efficiency. It can reduce the number of ML pipeline into one (at our problem setting), and the ML system effectively identifies OOD samples. Upon the trade-off between task accuracy and engineering efficiency, SOCAR adapted \textit{efficiency-centric} paradigm due to several aspects. First, the damage from faulty predictions at the image recognition task was not excessively toxic, as the professional human inspectors monitor the model's prediction results before business actions. Second, we prioritized robustness against OOD samples as we utilized user-generated images, which are hard to control. Lastly, we expect the \textit{efficiency-centric} ML system would evade the computation resource waste when there exist many tasks in the near future. To provide a guideline to the candidate ML practitioners, this section describes how we deployed our \textit{efficiency-centric} ML system in real world and how the end-users utilize it.

We deployed the proposed \textit{efficiency-centric} ML system in a cloud environment, especially under the Kubernetes Engine on the Google Cloud Platform. As shown in Figure \ref{fig:deploy_final}., the deployed system consists of serial actions of three components: producer pod, Kubernetes-based Event Driven Autoscaling (KEDA), and consumer pods. First, the producer pod retrieves the inference samples from the database, which is operated on the GCP BigQuery. When the cronjob operator triggers the operation, the producer publishes the inference dataset to the Cloud Pub/Sub. After the Cloud Pub/Sub receives the data from the producer pod, the KEDA \cite{keda} checks the number message in the Cloud Pub/Sub and decides the number of consumer pods established in the cloud environment. There were two primary drawbacks once we fixed the number of consumer pods in the prior deployment system. We experienced frequent overloads on consumer pods when many inference samples were retrieved. We also encountered a waste of resources on consumer pods when fewer inference samples existed. Therefore, we set the KEDA in our deployment system to efficiently manage the number of consumer pods based on the size of the inference dataset. Lastly, the consumer pods retrieve inference samples from the Cloud Pub/Sub, feed the sample into the \textit{efficiency-centric} OOD detector and the classifier to acquire the Top-2 prediction results. These results are inserted into the prediction result table, which is also managed with BigQuery. In a nutshell, we aim to design an efficient ML pipeline that can dynamically manage the number of consumer pods to maximize engineering efficiency. 


\section{Related Works}
\subsection{Multi-task Learning}
Multi-task learning \cite{zhang2021survey,crawshaw2020multi}, one of the traditional machine learning problems, aims to not only solve multiple tasks efficiently but also improve the performance of each task by exploiting the extracted knowledge from other tasks. There are well-known problems of training multiple tasks at once. First, when the scale of loss of each task varies, the model is easy to overfit to partial tasks. \cite{chen2018gradnorm} proposed to normalize gradient by using the norm of loss from each task to alleviate the problem. Second, knowledge from one task can negatively affect other tasks. \cite{lee2018deep} tries to prevent the negative transfer by introducing an asymmetric autoencoder term. The conventional architecture of multi-task learning in the vision domain comprises a single backbone network (i.e., CNN) and a few heads (linear layer) for each task. 

\subsection{Out-of-Distribution Detection}
The reliability of deep neural network is one of the most critical problems in real-world ML applications. The deployed model must be able to say `I don't know' when OOD input is given. Using pre-trained softmax deep neural network is one of the leading research directions. \cite{hendrycks2018baseline} gives a simple baseline, using the maximum value of softmax probability, for OOD detection. \cite{liang2020enhancing} shows temperature scaling and input pre-processing are effective for OOD detection. \cite{lee2018simple} proposes to use a simple Gaussian generative classifier and shows outstanding performance using Mahalanobis distance from test sample to mean of each class Gaussian. Even though many works based on the generative model to approximate training data distribution have emerged\cite{nalisnick2019deep, kirichenko2020normalizing}, logit-based methods show better performance. 

\section{Discussions and Conclusions}
In this study, we aim to introduce how we transformed prior task-centric ML system into the proposed efficiency-centric paradigm. We propose an efficient ML framework that concatenates multiple task-centric datasets, classifier, OOD detector, and prediction result table into a single pipeline to resolve this problem. We validated several takeaways throughout experiments on the various datasets retrieved from the real world car sharing platform. First, there exists a trade-off between task-centric and efficiency-centric paradigms. As the efficiency-centric ML system explores a larger label space, it acquired more general, less-biased representation power. We figured out this general representation power contributes to the competitive task accuracy with a supreme OOD detection performance. For the task-centric paradigm and other benchmarks, vice versa. Furthermore, we validated the efficiency-centric classifier calibrates better as it yields more qualified prediction results on the multi-labelable sample, while the task-centric classifier yields particularly biased prediction results. Lastly, we also elaborated on how we deployed this efficiency-centric ML system in the real world cloud environment. Still, there exist improvement avenues of our study. We would further excavate an underlying reason behind the trade-off between the task-centric and efficiency-centric paradigms. Moreover, we shall scrutinize how the efficiency-centric representation power differs from the baseline paradigms. Lastly, we would continuously evaluate the proposed paradigm's effectiveness and robustness under the increased number of tasks in the real world. As a closing remark, we highly expect ML practitioners to utilize the proposed efficiency-centric ML system in their domains for the sake of engineering efficiency in the real world.

\section{Acknowledgement}
We sincerely appreciate Sangtae Humphrey Ahn for his valuable guides, architecture designs, and supportings in deployment.

\newpage
\bibliographystyle{ACM-Reference-Format}
\bibliography{reference}



\newpage
\pagebreak


\end{document}